\documentclass[a4paper]{article}
\usepackage{INTERSPEECH2018}

\usepackage{amsmath}
\usepackage{multirow}
\usepackage{amssymb}
\usepackage{mathrsfs}
\usepackage{graphics}
\usepackage{graphicx}
\usepackage{ifsym}
\makeatletter
\g@addto@macro\normalsize{%
	\setlength\abovedisplayskip{0pt}
	\setlength\belowdisplayskip{0pt}
	\setlength\abovedisplayshortskip{0pt}
	\setlength\belowdisplayshortskip{0pt}
}
\makeatother
\usepackage{subcaption}
\usepackage{booktabs}
\usepackage{multirow}
\usepackage{array}
\usepackage{bbm}
\usepackage{color,colortbl}
\definecolor{Gray}{gray}{0.9}

\usepackage{multicol}
\usepackage{blindtext}
\usepackage[skip=1pt]{caption}
\usepackage{algorithm}
\usepackage{algpseudocode}
\usepackage{tikz}
\algnewcommand\algorithmicforeach{\textbf{for each}}
\algdef{S}[FOR]{ForEach}[1]{\algorithmicforeach\ #1\ \algorithmicdo}
\algnewcommand{\LineComment}[1]{\State \(\triangleright\) #1}

\algnewcommand\algorithmicinput{\textbf{Input:}}
\algnewcommand\INPUT{\item[\algorithmicinput]}
\algnewcommand\algorithmicoutput{\textbf{Output:}}
\algnewcommand\OUTPUT{\item[\algorithmicoutput]}

\algnewcommand\developericinput{\textbf{Software Developer Input:}}
\algnewcommand\DEVELOPERINPUT{\item[\developericinput]}

\algnewcommand\offlineicinput{\textbf{Offline Pre-trained Input:}}
\algnewcommand\OFFLINEINPUT{\item[\offlineicinput]}

\usepackage{hyperref}
\usepackage[nameinlink]{cleveref}
\hypersetup{
	colorlinks = true,
	linkcolor = red,
	citecolor = green
}
\usepackage{lipsum}
\usepackage{ntheorem}
\usepackage{makecell}

\renewcommand{\matrix}[1]{\mathbf{#1}}
\renewcommand{\vec}[1]{\mathbf{#1}}

\newcommand{\kmethod}{Prog-BiRNN\xspace}

\graphicspath {{figure/}}

\title{User Information Augmented Semantic Frame Parsing \\ using Coarse-to-Fine Neural Networks}
\name{Yilin Shen, Xiangyu Zeng, Yu Wang, Hongxia Jin}
\address{
  Samsung Research America, Mountain View, CA, USA
}
\email{\{yilin.shen,shane.z,yu.wang1,hongxia.jin\}@samsung.com}

\begin{document}

\maketitle

\begin{abstract}
  Semantic frame parsing is a crucial component in spoken language understanding (SLU) to build spoken dialog systems.
  It consists of two main tasks: intent detection and slot filling.
  State-of-the-art deep learning models have demonstrated good results on these tasks.
  However, these models require not only a large scale annotated training set but also a long training procedure.
  In this paper, we aim to alleviate these drawbacks for semantic frame parsing by utilizing the ubiquitous user information.
  We design a novel coarse-to-fine deep neural network model to incorporate prior knowledge of user information intermediately to better and quickly train a semantic frame parser.
  Due to the lack of benchmark dataset with real user information, we synthesize the simplest type of user information (location and time) on ATIS benchmark data.
  The results show that our approach leverages such simple user information to outperform state-of-the-art approaches by 0.25\% for intent detection and 0.31\% for slot filling using standard training data.
  When using smaller training data, the performance improvement on intent detection and slot filling reaches up to 1.35\% and 1.20\% respectively.
  We also show that our approach can achieve similar performance as state-of-the-art approaches by using less than 80\% annotated training data.
  Moreover, the training time to achieve the similar performance is also reduced by over 60\%.
\end{abstract}

\vspace{2pt}

\section{Introduction}

With the emergence of artificially intelligent voice-enabled personal assistants in daily life, spoken language understanding (SLU) system has attracted increasing research attentions.
As the key component in a SLU system, semantic frame parsing aims to identify user's intent and extract semantic constituents from a natural language utterance, a.k.a. intent detection and slot filling.
Existing approaches includes the independent models for learning intent detection \cite{haffner2003optimizing, liu2016attention} and slot filling \cite{McCallum2000MEM, Yao2014slottagging, mesnil2015using, peng2015recurrent, liu2015recurrent, kurata2016leveraging} separately as well as joint models to learn these two tasks together \cite{guo2014joint,Xu2013joint,hakkani2016multi,liu2016attention}.

Unfortunately, the aforementioned approaches suffer from several main drawbacks.
First, they require the existence of a large scale annotated corpus to train a high quality parser.
Since a SLU system aims to understand all varieties of user utterances, the corpus is further required to extensively cover all varieties of utterances.
However, the collection of such an annotated corpus is very expensive and needs heavy human labor.
Secondly, the training of existing parser models oftentimes takes a long time to achieve a good performance.
These drawbacks are magnified especially with the recent quick growth of capabilities in personal assistants \cite{skill_statistics}.
To develop a new domain, we need to generate a new utterance dataset and take a long time to train a new semantic frame parsing model.
Thus, it is critically desirable to design a new semantic frame parsing model to alleviate the needs of both large amount of annotated training data and long training time.

In this paper, we investigate how user information can be incorporated into semantic frame parsing to overcome the above drawbacks.
We design a novel progressive attention-based recurrent neural network (\kmethod) model that first annotates the information types and then distills the related prior knowledge w.r.t. each type of information to continue learning intent detection and slot filling.
Our approach is motivated by the recent success of attention-based RNN model \cite{liu2016attention} for joint learning of intent detection and slot filling and coarse-to-fine neural networks \cite{Rusu2016progressive} in many multi-tasking learning applications.
Our model includes a main RNN structure stacked with a set of different layers and they are trained one by one in a progressive manner.

\emph{Organization:}
Section \ref{sc:background} describes the background and related work.
We discuss our new problem definition in Section \ref{sc:problem}.
Section \ref{sc:approach} includes our proposed model and its training procedure details.
We show the experimental results in Section \ref{sc:experiment}.
Section \ref{sc:discussion} concludes the whole paper.

\section{Background \& Related Work}\label{sc:background}

\subsection{Semantic Frame Parsing}\label{sc:related}

Intent detection and slot filling are two main tasks to build a semantic frame parser for spoken language understanding (SLU).
That is, the goal of semantic frame parsing is to understand all varieties of user utterances by correctly identifying user's intents and slot tags.
Given an input utterance as a sequence $\vec{x}$ of length $T$, intent detection identifies the intent class $I$ for $\vec{x}$ and slot filling maps $\vec{x}$ to the corresponding label sequence $\vec{y}$ of the same length $T$ (Table \ref{table:atis_example}).

Intent detection is treated as an utterance classification problem, which can be modeled using conventional classifiers such as support vector machine (SVM) \cite{haffner2003optimizing} and RNN based models \cite{liu2016attention}.
As a sequence labeling problem, slot filling can be solved using traditional machine learning approaches including maximum entropy Markov model \cite{McCallum2000MEM} and conditional random fields (CRF) \cite{Raymond2007GenerativeAD}, as well as recurrent neural network (RNN) based approaches which takes and tags each word in an utterance one by one \cite{Yao2014slottagging,mesnil2015using,peng2015recurrent,liu2015recurrent,kurata2016leveraging}.
Recent research focuses on the joint model to learn two tasks together \cite{guo2014joint,Xu2013joint,hakkani2016multi,liu2016attention}.

\subsection{Joint Attention-based RNN Model}

We recall the state-of-the-art approach in \cite{liu2016attention}, referred to as \emph{Att-BiRNN} model, which will be used as the base of our approach.
Att-BiRNN is a joint RNN model to learn the two tasks together.
It first uses a bidirectional RNN with a basic LSTM cell to read the input utterance as a sequence $\vec{x}$.
At each time stamp $t$, a context vector $\vec{c}_t$ is learned to concatenate with the RNN hidden state $\vec{h}_t$, i.e., $\vec{c}_t \oplus \vec{h}_t$, to learn a slot attention for predicting the slot tag $y_t$.
All hidden states of slot filling attention layer are used to predict the intent label in the end.
The objective function of Att-BiRNN model is as follows:
\begin{equation}\label{eq:rnn_obj}
P(\vec{y}|\vec{x})= \max_{\theta_r, \theta_s, \theta_I} \prod_{t=1}^T P(y_t | y_1, \ldots, y_{t-1}, \vec{x}; \theta_r, \theta_s, \theta_I)
\end{equation}
where $\theta_r, \theta_s, \theta_I$ are the trainable parameters of different components (utterance BiRNN, slot filling attention layer and intent classifier) in Att-BiRNN model.

%RNN encoder-decoder framework reads the input sequence $\vec{x}$ using one encoder RNN to derive the hidden state of the encoder $h_{e}$.
%The decoder RNN then uses $h_{e}$ to generate the target output sequence $\vec{y}$, with the following objective function:
%\begin{equation}\label{eq:rnn_enc_dec_obj}
%P(\vec{y})= \max_{\theta} \prod_{t=1}^T P(y_t | y_1, \ldots, y_{t-1}, h_{e}; \theta)
%\end{equation}
%which depends on the hidden state $c$ from encoder RNN rather than input sequence $\vec{x}$ in equation \eqref{eq:rnn_obj}.
%%However, neither RNN or RNN encoder-decoder model can take external knowledge with different semantic meanings.

\section{Problem Definition}\label{sc:problem}

%The existing approaches suffer from several limitations.
%First, they heavily rely on a training set, $\{(\vec{x}^{(n)}, \vec{y}^{(n)}): n=1,\ldots,N\}$, to learn a high quality semantic frame parser.
%That said, there must exist a training set which contains a large number of sentences with all varieties.
%Unfortunately, it is costly and time consuming to collect such a \emph{ideal} training set.
%Secondly, the input utterance itself does not provide enough information in order to correctly parse its semantic frame.
%As the example in Table \ref{table:atis_example} shows, \textsl{``LA"} could be either ``fromloc" or ``toloc" in the context of this utterance and similar as \textsl{``Boston"}.

We propose the \emph{User Info Augmented Semantic Frame Parsing} problem for the same two tasks, intent detection and slot filling, by considering the following additional inputs.

\vspace{2pt}
\noindent\emph{User Info Dictionary:}
This defines the categorical relations between user info type and slots.
In other words, each key in the dictionary is a type of user info and its corresponding value is the slots belonging to this type.
%Examples in ATIS data is shown in Table \ref{table:user_info_dict}.
The generation of this dictionary is not the focus of our paper since it can be simply generated by a software developer when he generates slots during the development of a new domain in practice.

Each type of user info is associated with an external or pre-trained model to extract their semantically meaningful prior knowledge.
For example, the semantics of a location is represented by its longitude and latitude such that the distance between two locations reflect their actual geographical distance.

\vspace{2pt}
\noindent\emph{User Info for Each Utterance:}
Each input sequence $\vec{x}$ is associated with its corresponding user info $U$.
$U$ is represented as a set of tuples, $\langle \textsl{Info Type}, \textsl{Info Content} \rangle$.
As an example utterance in Table \ref{table:atis_example}, the first gray row shows our generated user info with type $\textsl{``User Location"}$ and content $\textsl{``Brooklyn, NY"}$. 
Learning user info has been well studied, such as user contextual information (e.g., time, location, activity, etc.) via smartphone \cite{Yurur2016MobileSurvey}, Internet of Things \cite{Perera2014IoTSurvey} and user interests (e.g., favorite food, etc.) using recommendation models \cite{Su2009SCF}.

%Due to the lack of real user info for the widely used benchmark datasets in SLU research, we will introduce later in the experiment about how we synthesize user info dictionary and user info for each utterance in ATIS benchmark dataset.

%At last, our goal is to achieve both intent detection and slot filling tasks with better performance.

\vspace{2pt}
\noindent\textbf{\em Remarks:}
One may argue that this is a simple extension of semantic frame parsing problem in which the user info can be simply encoded into an existing model as a new input or a new state. 
However, these naive approaches ignore the different semantic meanings between user info and language context in an utterance, as well as between different types of user info.
Thus, as we later show in experiment (Section \ref{sc:experiment}), these baseline approaches do not show any advantage over existing approaches without user info.

\section{Proposed Approach}\label{sc:approach}

In this section, we describe the main idea and details of our proposed \kmethod model as well as its training procedure.

%\subsection{Progressive Neural Networks Models}
\subsection{Coarse-to-Fine Attention-based RNN Model}

As the name indicates, our main idea is to train the semantic frame parsing model from coarse to fine progressively with an intermediate task before achieving the final goal of intent detection and slot filling.
This is motivated by the recent success of progressive neural networks \cite{Rusu2016progressive}.
Specifically, for each utterance $\vec{x}$, we first define the user info sequence $\vec{z}$ using the user info dictionary.
In Table \ref{table:atis_example}, the last row shows the user info sequence corresponding to this example.
Our approach first trains a user info tagging to derive $\vec{z}$.
Then, the prior knowledge with semantic meaning for each type of user info is distilled into the model to continue training for intent detection and slot filling.

\setlength{\tabcolsep}{0.05em}
\begin{table}[h]\scriptsize
	\centering
	\caption{ATIS corpus sample with intent and slot annotations with additional user info and its corresponding user info sequence (in gray)}
	\label{table:atis_example}
	\begin{tabular}{c|c|c|c|c|c|c|c}
		\toprule
		\textbf{utterance} ($\vec{x}$) & round & trip & flights & between & ny & and & miami \\
		\midrule
		\textbf{slots} ($\vec{y}$) & B-round\_trip & I-round\_trip & O & O & B-fromloc & O & B-toloc \\
		\midrule
		\textbf{intent} ($I$) &  \multicolumn{7}{c}{atis\_flight} \\
		\midrule
		\midrule
		\rowcolor{Gray}
		\textbf{user info} ($U$) &  \multicolumn{7}{c}{$\{\textsl{``User Location"}:\textsl{``Brooklyn, NY"}\}$} \\
		\midrule
		\rowcolor{Gray}
		\textbf{user info seq} ($\vec{z}$) & O & O & O & O & B-loc & O & B-loc \\
		\bottomrule		
	\end{tabular}
\end{table}

%Next, we first present our proposed models with only B- and O- format for the sake of clear presentation.
%We then discuss the progressive training algorithm and how IOB format is supported in our models.

%\subsubsection{Progressive Attention-based RNN Model}

\begin{figure}[t]
	\centering
	\includegraphics[width=0.9\columnwidth,height=60mm]{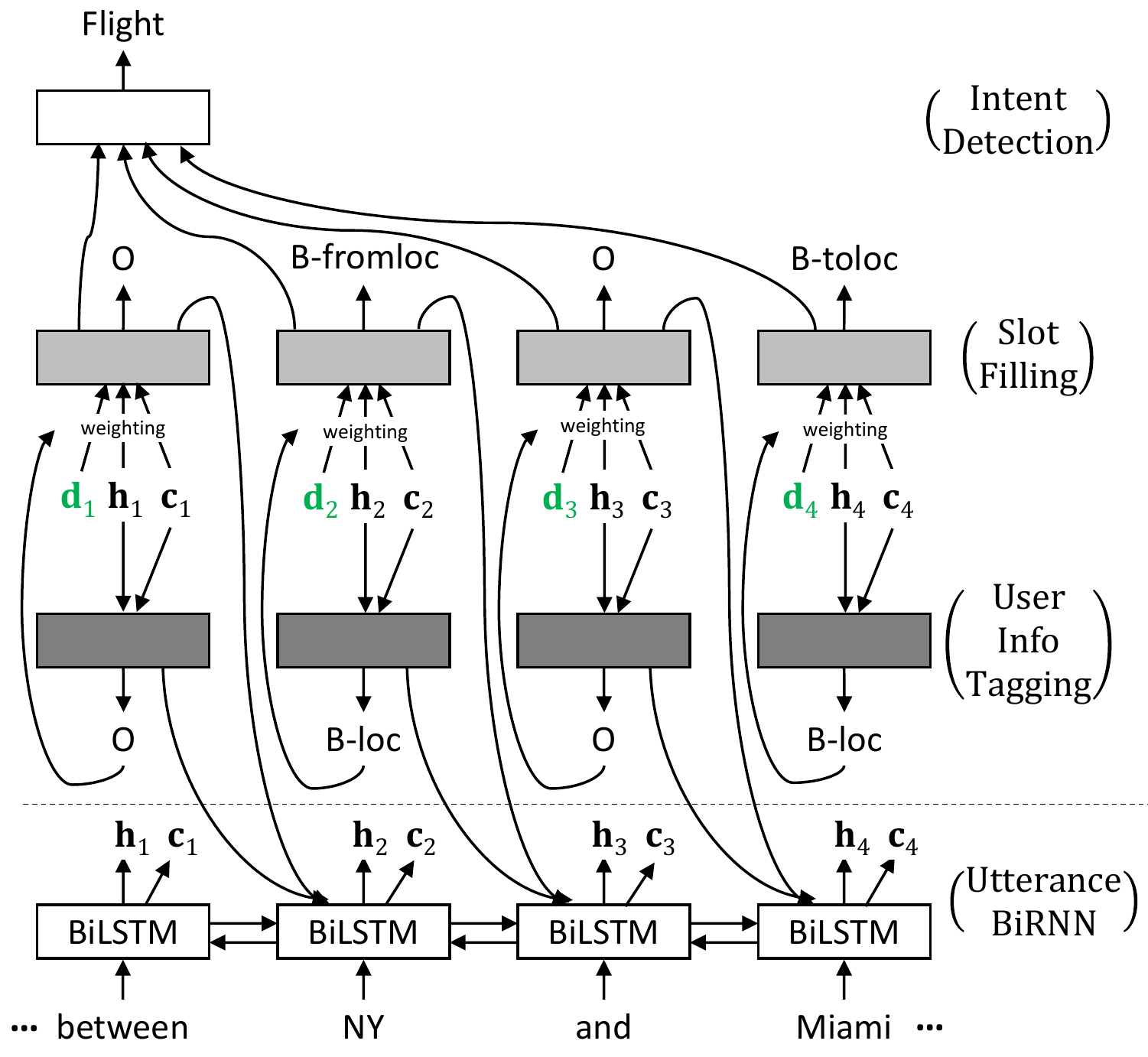}
	\caption{Coarse-to-Fine Attention based RNN Model}
	\label{fig:attention_rnn}
\end{figure}

As shown in Figure \ref{fig:attention_rnn}, our proposed \kmethod model is designed based on the state-of-the-art Att-BiRNN model \cite{liu2016attention}, which consists of the following four main components.

\vspace{2pt}
\noindent\textbf{Utterance BiRNN Layer:}
We use the same bidirectional RNN (BiRNN) to encode an utterance with LSTM cells (BiLSTM) as in \cite{liu2016attention}.
The hidden state $\vec{h}_t$ at each time step $t$ is the concatenation of forward state $\vec{fh}_t$ and backward state $\vec{bh}_t$, i.e., $\vec{h}_t = \vec{fh}_t \oplus \vec{bh}_t$.

\vspace{2pt}
\noindent\textbf{User Info Tagging Layer:}
This component labels the user info type for each word in the input utterance.
Since the labeling is based on the language context of input utterance, we follow the previous work \cite{liu2016attention} to use a language context vector $\vec{c}_t$ at each time stamp $t$ via the weighted sum of all hidden states $\{\vec{h}_k\}_{\forall 1\le k \le T}$ i.e., $\vec{c}_t = \sum_{k=1}^T \alpha_{t,k} \vec{h}_k$.
Here, $\pmb{\alpha}_t = \textsl{softmax}(\vec{e}_t)$, i.e., $\alpha_{t,j} = {\exp(e_{t,j}) \over \sum_{k=1}^T \exp(e_{t,k})}$.
$e_{t,k} = g(\vec{s}_{t-1}^u, \vec{h}_k)$ is also learned from a feed forward neural network $g$ with the previous hidden state $\vec{s}_{t-1}^u$ defined as the concatenation of $\vec{h}_{t-1}$ and $\vec{c}_{t-1}$, i.e., $\vec{s}_{t-1}^u = \vec{h}_{t-1} \oplus \vec{c}_{t-1}$.
At each time step $t$, the user info tagging layer outputs $P_u(t)$ as follows:
\begin{equation}
P^u_t = \textsl{softmax} (\matrix{W}^u \vec{s}_t^u);\;\; \tilde{z}_t = \arg\max_{\theta_u} P^u_t
\end{equation}

\vspace{2pt}
\noindent\textbf{Slot Filling Layer:}
This is the key layer for distilling user info into the model to help reduce the need of annotated training data.
It shares the same hidden state $\vec{h}_t$ and language context $\vec{c}_t$ with the user info tagging layer.
%We next discuss how to distill the external prior knowledge from user info.
For each word in the utterance, we use external knowledge to derive the prior distance vectors $\vec{d}_t = \{\vec{d}_t(1), \ldots, \vec{d}_t(|U|)\}$ for each time stamp $t$ (green in Figure \ref{fig:attention_rnn}) where $|U|$ is the number of user info types in IOB format.
And each element $\vec{d}_t^j$ is defined as follows:
\begin{equation}
\vec{d}_t(j) = \textsl{sigmoid} \Big( \pmb{\beta}(j) \odot \pmb{\delta}_t(j) \Big)
\end{equation}
where $\odot$ stands for element-wise multiplication.
$\pmb{\beta}(j)$ is a $|U|$ dimensional trainable vector; and $\pmb{\delta}_t(j)$ is the distance between the $t^\textsl{th}$ word and user info w.r.t. the prior knowledge of type $j$.

Next, we define the calculation of distance $\delta_t(j)$ for each info type $j$ at time stamp $t$, through the example in Figure \ref{fig:attention_rnn}.
Let $\delta_t(\textsl{loc})$ be the distance w.r.t. the location type of user info.
It is a one-dimensional scalar in this case.
Taking the second word \textsl{``NY"} as an example, we have its following location distance since it is tagged as \textsl{``Location"} type of user info:
$$\delta_2(\textsl{loc}) = \textsl{dist}\textsl{(``NY", ``Brooklyn, NY")} \approx \textsl{4.8 (miles)}$$ by using external location based services, i.e., Google Maps Distance Matrix API \cite{googledist}.
If the word and user info are of different types, we set the distance $\delta_t(j)$ as -1 such that its corresponding $d_t(j)$ will be close to 0 via the sigmoid function.

To feed the prior distance vectors $\vec{d}_t$ into the slot filling layer, we weight each element $\vec{d}_t(j)$ and the language context $\vec{c}_t$ over the softmax probability distribution $P^u_t$ from the user info tagging layer.
Intuitively, this determines how important a type of user info or the language context in utterance is to predict the slot tag of each word in the utterance.
Thus, we have the input $\pmb{\Phi}_t$ of LSTM cell at each time step $t$ in slot filling layer as follows:
\begin{equation}\label{eq:phi}
\pmb{\Phi}_t = P_t^u(1) \vec{d}_t(1) \oplus \cdots \oplus P_t^u(|U|) \vec{d}_t(|U|)  \oplus P_t^O  \vec{c}_t
\end{equation}
where $P^u_t(j)$ and $P_t^O$ stand for the probability that the $t^\textsl{th}$ word is predicted as $j$ type of user info and as ``O" meaning none of the types.
\emph{Note that we will discuss how to deal with IOB format in Section \ref{sc:approach:iob}.}
At last, the state $\vec{s}_t^s$ at time step $t$ is computed as $\vec{h}_t \oplus \pmb{\Phi}_t$ and the slot tag is predicted as follows:
\begin{equation}
P_t^s = \matrix{W}^s \vec{s}_t^s;\;\; \tilde{y}_t = \arg\max_{\theta_s} P^s_t
\end{equation}

\vspace{2pt}
\noindent\textbf{Intent Detection Layer:}
We add an additional intent detection layer as in \cite{liu2016attention} to generate the probability distribution $P_I$ of intent class labels by using the concatenation of hidden states from slot filling layer, i.e., $\vec{s}^I = \vec{s}_1^s \oplus \ldots \oplus \vec{s}_T^s$.
\begin{equation*}
P^I = \textsl{softmax}(\matrix{W}^I \vec{s}^I);\;\; \tilde{I} = \arg\max_{\theta_I} P^I
\end{equation*}

\noindent\textbf{\em Remarks:}
The sharing of hidden state $\vec{h}_t$ and language context $\vec{c}_t$ between user info tagging and slot filling layers is crucial to reduced the required annotated training data.
For the user info tagging layer, $\vec{h}_t, \vec{c}_t$ are mainly used to tag the words which belong to one type of user info.
The semantic slots of these words can be easily tagged in slot filling layer by utilizing the distilled prior knowledge instead of using $\vec{h}_t, \vec{c}_t$ again.
The slot filling then depends on $\vec{h}_t, \vec{c}_t$ to tag the rest of words not belonging to any type of user info.

\subsection{Progressive Training with IOB Format Support}\label{sc:approach:training}

\subsubsection{Training Algorithm}

The training procedure is progressively conducted step by step.
The first step is to train user info tagging component with loss function $\mathcal{L}_u$ as follows:
\begin{equation}
\mathcal{L}_u (\theta_r, \theta_u) \triangleq - {1 \over n} \sum_{i=1}^{|U|} \sum_{t=1}^n z_t(i) \log P_t^u(i)
\end{equation}
where $|U|$ is the number of user info types in IOB format.

Then, we train the slot filling layer with loss function $\mathcal{L}_s$ and intent classifier with loss function $\mathcal{L}_I$ simultaneously.
In the meanwhile, we also allow the fine tuning of parameters $\theta_r$ and $\theta_u$ in utterance BiRNN and user info tagging layers.
\begin{equation}
\mathcal{L}_s (\theta_r, \theta_I, \theta_s, \theta_u) \triangleq - {1 \over n} \sum_{i=1}^{|S|} \sum_{t=1}^n y_t(i) \log P_t^s(i)
\end{equation}

\begin{equation}
\mathcal{L}_I (\theta_r, \theta_I, \theta_s, \theta_u) \triangleq - \sum_{i=1}^{|I|} I(i) \log P^I(i)
\end{equation}
where $|S|$ is the number of slots in IOB format and $|I|$ is the number of intents.
$P(i)$ stands for the probability $P(X=x_i)$.
Moreover, $\theta_r, \theta_u, \theta_s, \theta_I$ are the parameters in utterance BiRNN, user info tagging, slot filling and intent detection components in our proposed \kmethod model.

\subsubsection{Details of IOB Format Support}\label{sc:approach:iob}

\begin{figure}[h]
	\centering
	\includegraphics[width=0.75\columnwidth,height=30mm]{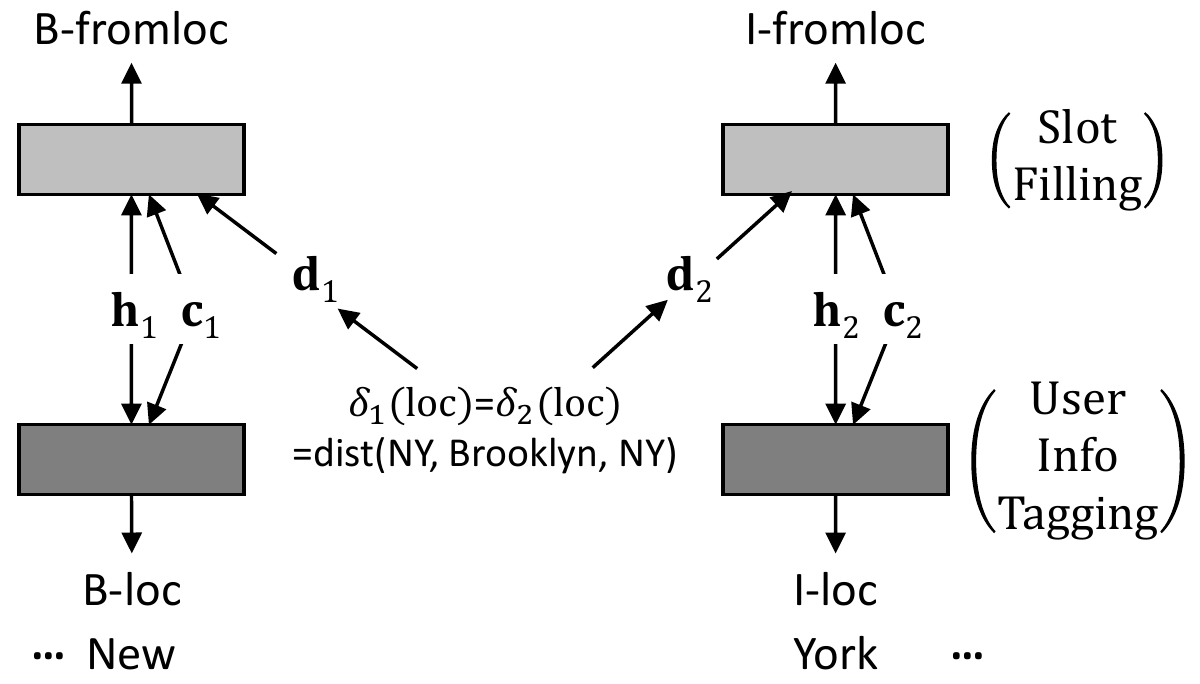}
	\caption{Support of IOB Format (omitted other model details)}
	\label{fig:iob}
\end{figure}

Thanks to the progressive training procedure, the IOB format will be naturally supported in our model.
As shown in Figure \ref{fig:iob}, in the case of \textsl{``New York"} with ``B-loc I-loc" user info tags, we take them together to extract the prior geographical distance $\textsl{dist}\textsl{(``New York", ``Brooklyn, NY")}$.
Moreover, since B-loc and I-loc are considered as different tags in the output $P_t^u$ of user info tagging component, they can be directly used to infer B-fromloc and I-fromloc in slot filling component accordingly.

In the case that the type of user info for the $t^\textsl{th}$ word is incorrectly tagged, the hidden state $\vec{h}_t$ and language context $\vec{c}_t$ will be used to infer the slot tags since the user info tagging output $P_t^u$ will weight more on $\vec{h}_t, \vec{c}_t$ in this case.
In addition, the second phase of training procedure for joint training of all components also leans to use more language context to correct the incorrectly tagged type of user info.

\noindent\textbf{\em Remarks:}
The capability of prior knowledge distillation in our approach leverages user information to largely improve the performance and reduce the requirement of annotated training data.
Moreover, the overall training time is also largely shortened since our approach divides SLU into simpler subproblems in which each subproblem is much easier to train.

\section{Experimental Evaluation}\label{sc:experiment}

\subsection{Dataset}

We evaluate our approach on the ATIS (Airline Travel Information Systems) dataset \cite{Hemphill1990ATIS}, a widely used dataset in SLU research.
The training set contains 4,978 utterances from the ATIS-2 and ATIS-3 corpora, and the test set contains 893 utterances from the ATIS-3 data sets.
There are 127 distinct slot labels and 22 different intent classes.

Due to the lack of benchmark datasets with user info, we design the following two mechanisms to synthesize two types of user info, \emph{user contextual location} and \emph{user preferred time periods} in ATIS dataset.
%\footnote{\scriptsize{We will release the dataset upon publication.}}.
%As shown in Table \ref{table:user_info_dict}, 
We first construct the user info dictionary by including all slots with "loc" keyword in contextual location and including all slots with "time" keyword in user preferred time period.
%Thus, the user info sequence $\vec{z}$ can be generated by replacing the slots in Table \ref{table:user_info_dict} with its corresponding user info in IOB format, and other slots with \textsl{'O'}.

%\setlength{\tabcolsep}{0.5em}
%\begin{table}[h]\scriptsize
%	\centering
%	\caption{User info dictionary for ATIS dataset}
%	\label{table:user_info_dict}
%	\begin{tabular}{p{1.6cm}|p{5.8cm}}
%		\toprule
%		\textbf{User Info Type} & \multicolumn{1}{c}{\textbf{Slots}} \\
%		\midrule
%		\textbf{Contextual Location} &  toloc.airport\_code, toloc.state\_code, fromloc.state\_name, toloc.state\_name, toloc.city\_name, fromloc.city\_name, depart\_time.start\_time, depart\_time.end\_time, fromloc.state\_code, arrive\_time.time, depart\_time.time, toloc.country\_name, stoploc.airport\_name, arrive\_time.end\_time, toloc.airport\_name, stoploc.state\_code, fromloc.airport\_code, fromloc.airport\_name, stoploc.city\_name \\
%		\midrule
%		\textbf{Preferred Time Periods} &  depart\_time.period\_of\_day, time, depart\_time.start\_time, depart\_time.end\_time, arrive\_time.time, depart\_time.time, arrive\_time.start\_time, arrive\_time.period\_of\_day, arrive\_time.end\_time, return\_time.period\_of\_day \\
%		\bottomrule		
%	\end{tabular}
%\end{table}

The prior distance $\pmb{\delta}$ of contextual location are computed using Google Maps Distance Matrix API \cite{googledist}.
For time period, we calculate $\pmb{\delta}$ by using the difference between the tagged time stamp in an utterance and the middle time stamp of the user preferred time period.

\vspace{2pt}
\noindent\emph{Contextual Location:}
W.l.o.g., we synthesize user contextual locations based on the intuitive assumption that user's location is usually close to flight depart city.
We first extract all values (real locations) of slots which contains "fromloc" in their names.
Then, for each real location, we use Google Places API \cite{googleplace} to find the nearby cities within 50 km.
For each utterance having slots containing 'fromloc', we add the nearby city of this slot value as its location.
When there are more than one nearby cities, we randomly select one from them.

\vspace{2pt}
\noindent\emph{Preferred Time Periods:}
We follow Oxford dictionary to consider four periods of a day: morning (6am-12pm), afternoon (12pm-6pm), evening (6pm-12am), night (12am-6am).
In each utterance having the slots with "time" keyword, we generate one depart and one arrive time preference by selecting from these four periods as follows:
If there is a slot containing 'depart\_time', we set the preferred time period based on the value of this slot.
For example, if the slot value is ``8pm", we set the preferred time period to be ``evening" since ``8pm" belongs to the period 6pm-12am.
For the slot 'depart\_time.period\_of\_day', we simply match the key words to synthesize the user preferred depart time period.
We synthesize the arrive period preference in the same way.

\setlength{\tabcolsep}{0.2em}
\begin{table}[t]\scriptsize
	\centering
	\caption{Examples of synthesized user info in ATIS dataset}
	\label{table:sample_data}
	\begin{tabular}{c|c|c}
		\toprule
		\multirow{2}{*}{\textbf{Utterance}} & \multicolumn{2}{c}{\textbf{User Info}}\\
		& Type & Content \\
		\midrule
		\makecell{i need a flight from dallas to san francisco \\ \{``fromloc.city\_name": ``dallas"\}} & contextual location & Fort Worth,TX \\ 
		\midrule
		\makecell{all flights to baltimore after 6 pm \\ \{``depart\_time.time": ``6 pm"\}} & \makecell{preferred\\depart period} & evening \\
		\midrule
		\makecell{i want to fly from boston at 838 am and \\ arrive in denver at 1110 in the morning \\ \{``fromloc.city\_name": ``boston"\} \\ \{``arrive\_time.time": ``1110"\} \\ \{``arrive\_time.period\_of\_day": ``morning"\}
		}
		& \makecell{contextual location \\ \\ \makecell{preferred\\arrive period}} & \makecell{Cambridge,MA \\ \\ morning}\\
		\bottomrule
	\end{tabular}
\end{table}

\subsection{Baseline Competitors \& Implementation Details}
In addition to the state-of-the-art baseline Att-BiRNN in \cite{liu2016attention}, we also design another baseline competitor using user info as discussed at the end of Section \ref{sc:problem}.
For the sake of fairness, we consider concatenating the user info directly to the input of slot filling layer in the Att-BiRNN.
All user info is concatenated together without distinguishing different types.
We call these two baselines \emph{Att-BiRNN with/without User Info} respectively.

Also, we follow the exact same hyperparameters in the original paper of the base Att-BiRNN model \cite{liu2016attention} since our model does not have additional hyperparameters.

\subsection{Results with Different Sizes of Training Set}

\begin{figure}[t]
	\centering
	\begin{subfigure}{0.49\textwidth}
		\centering
		\includegraphics[width=0.45\textwidth]{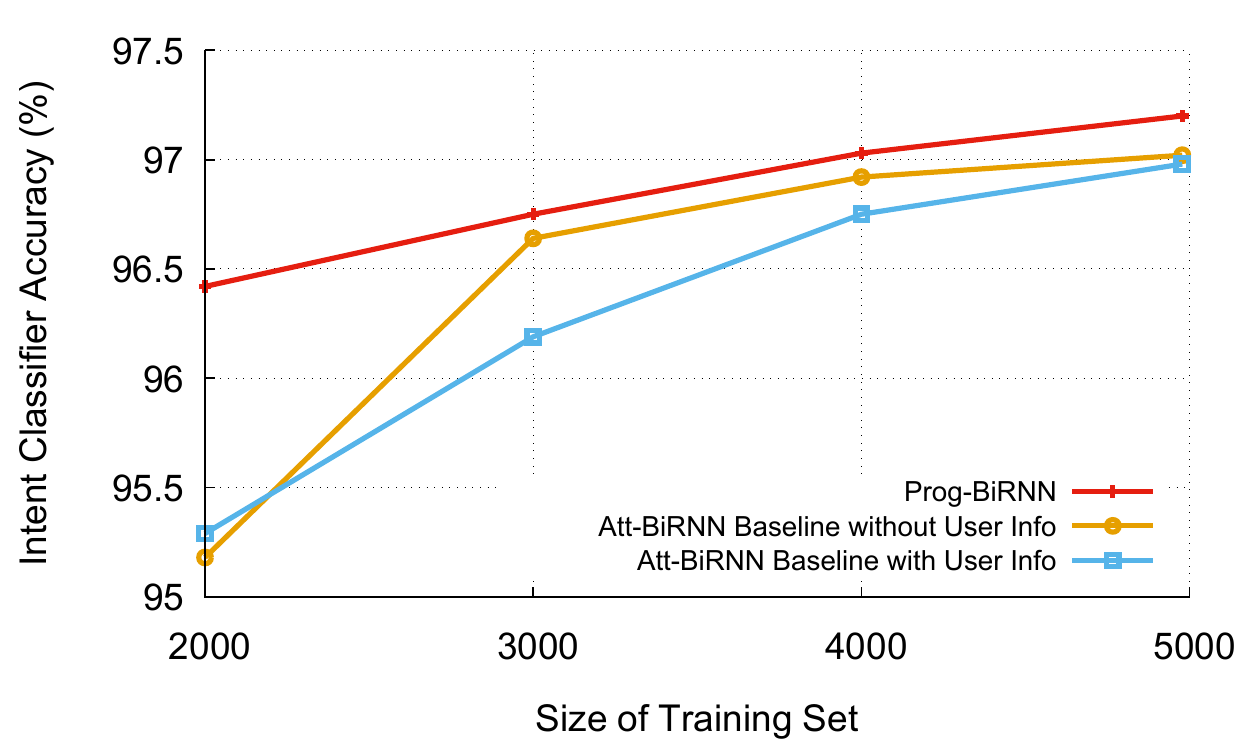}
		\includegraphics[width=0.45\textwidth]{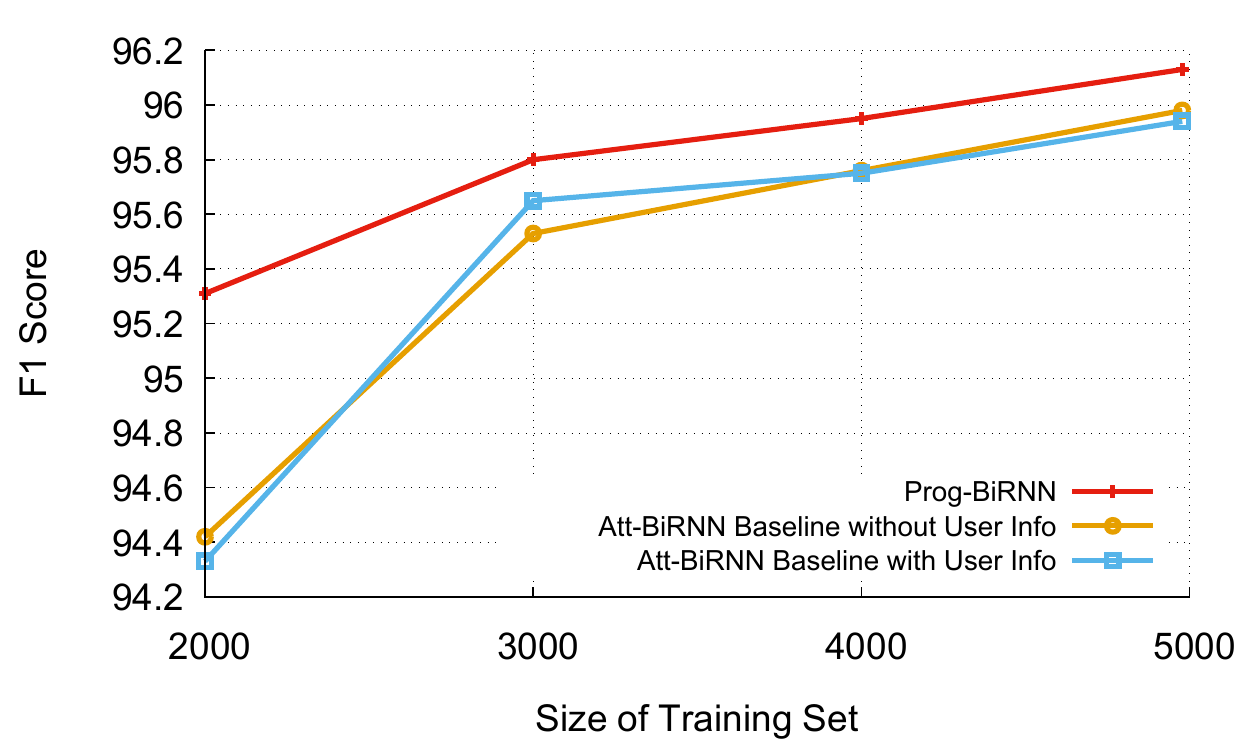}
		\caption{Contextual Location Only}
		\label{fig:performance:location}
	\end{subfigure}
	\begin{subfigure}{0.49\textwidth}
		\centering
		\includegraphics[width=0.45\textwidth]{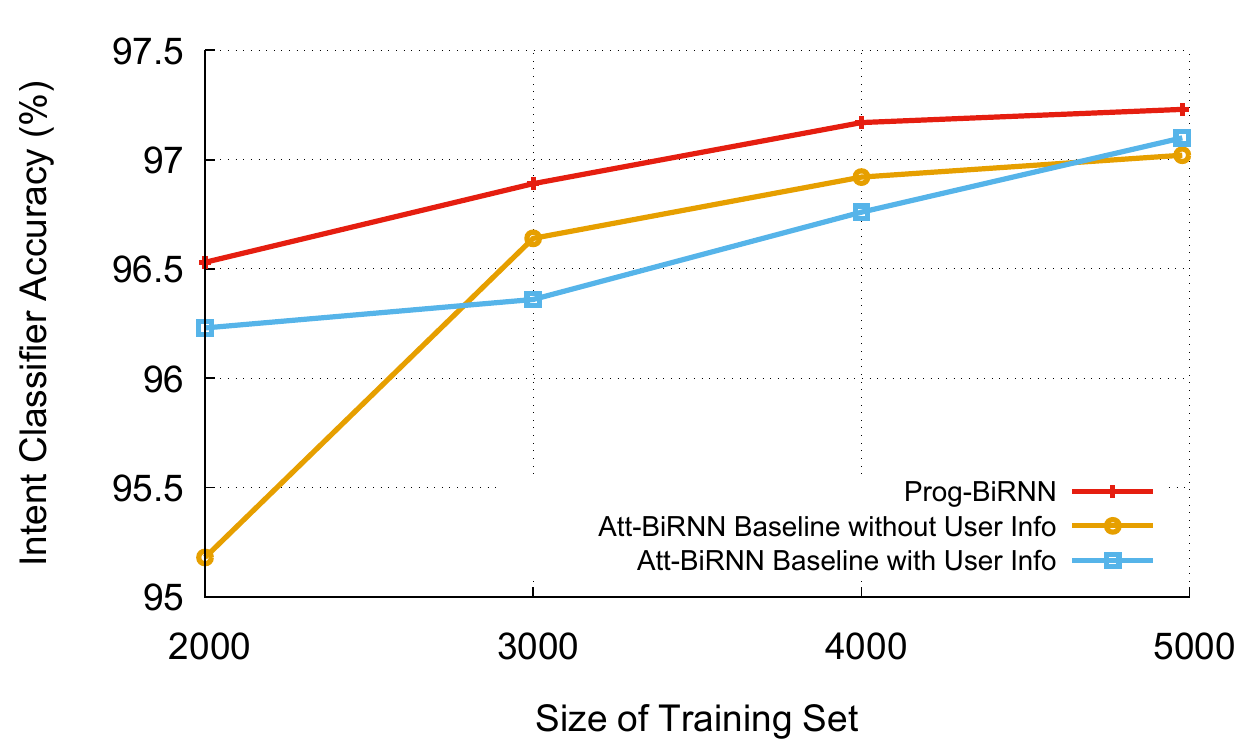}
		\includegraphics[width=0.45\textwidth]{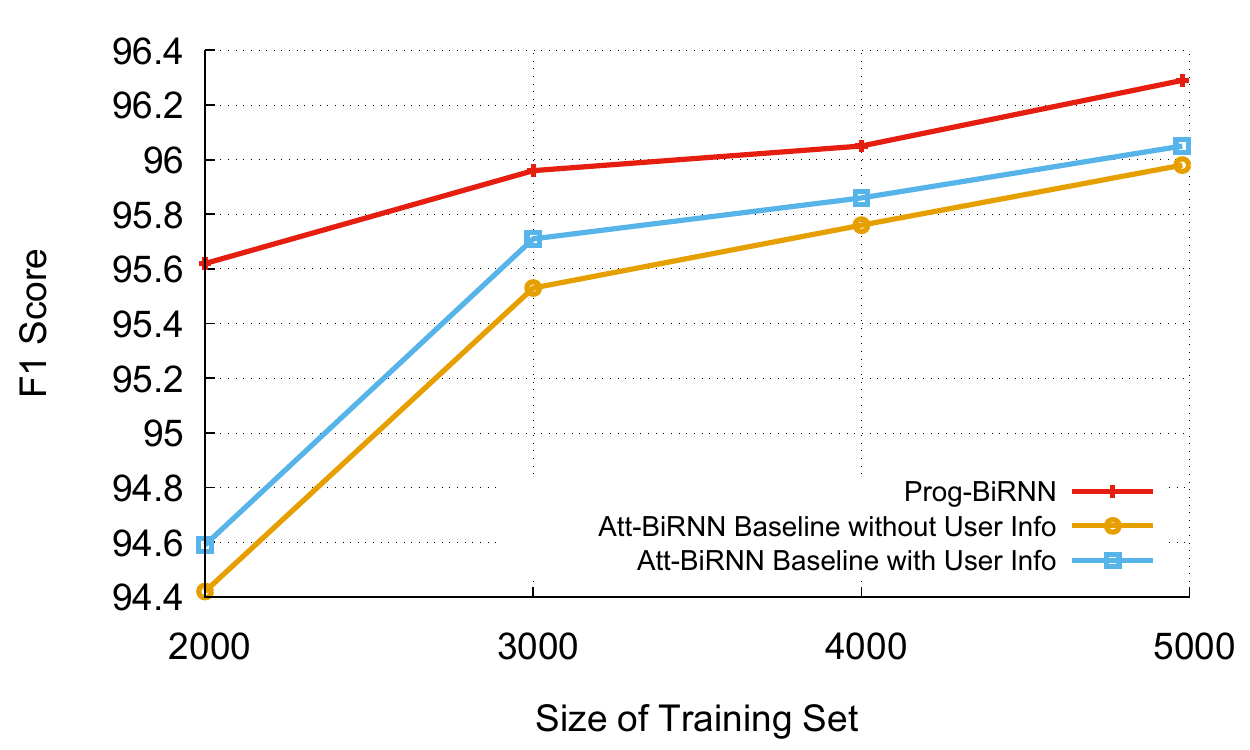}
		\caption{Contextual Location \& Preferred Time Periods}
		\label{fig:performance:locationtime}
	\end{subfigure}
	\caption{Performance results with different sizes of training set}
	\label{fig:performance}
\end{figure}

We evaluate our \kmethod model on subsets of full size ATIS training set and randomly sampled 3 different sizes (2,000, 3,000 and 4,000) utterances out of the total 4,978 utterances.
Figure \ref{fig:performance} reports the average performance results on 10 differently sampled training set of each size.

Since location related slots are the majority of all slots in ATIS dataset, we first consider only using contextual location as user info.
As shown in Figure \ref{fig:performance:location}, the F1 score of slot filling outperforms both baseline approaches with around 0.2\% absolute gain of each size.
The accuracy improvement of intent detection is around 0.1\% and up to 0.2\% for full size training set.
This slightly smaller improvement margin is due to the small number of intent classes.
When using both contextual location and preferred time period as user info, we observe more significant improvement with 0.25\% gain for intent detection and 0.31\% gain for slot filling.
Note that our reported intent detection accuracy is different from that in baseline paper \cite{liu2016attention} since we use all 22 intents in ATIS dataset.
In particular, when using smaller training data, i.e., 2000 training data, the performance improvement on intent detection and slot filling reaches 1.35\% and 1.20\% respectively.
More significantly, our \kmethod model can use less than 4000 (80\%) annotated utterances with simple user location and preferred time period as training data to achieve the performance of baseline approaches for both intent detection and slot filling.

\subsection{Training Time Results}

\begin{figure}[h]
	\centering
	\centering
	\includegraphics[width=0.29\textwidth]{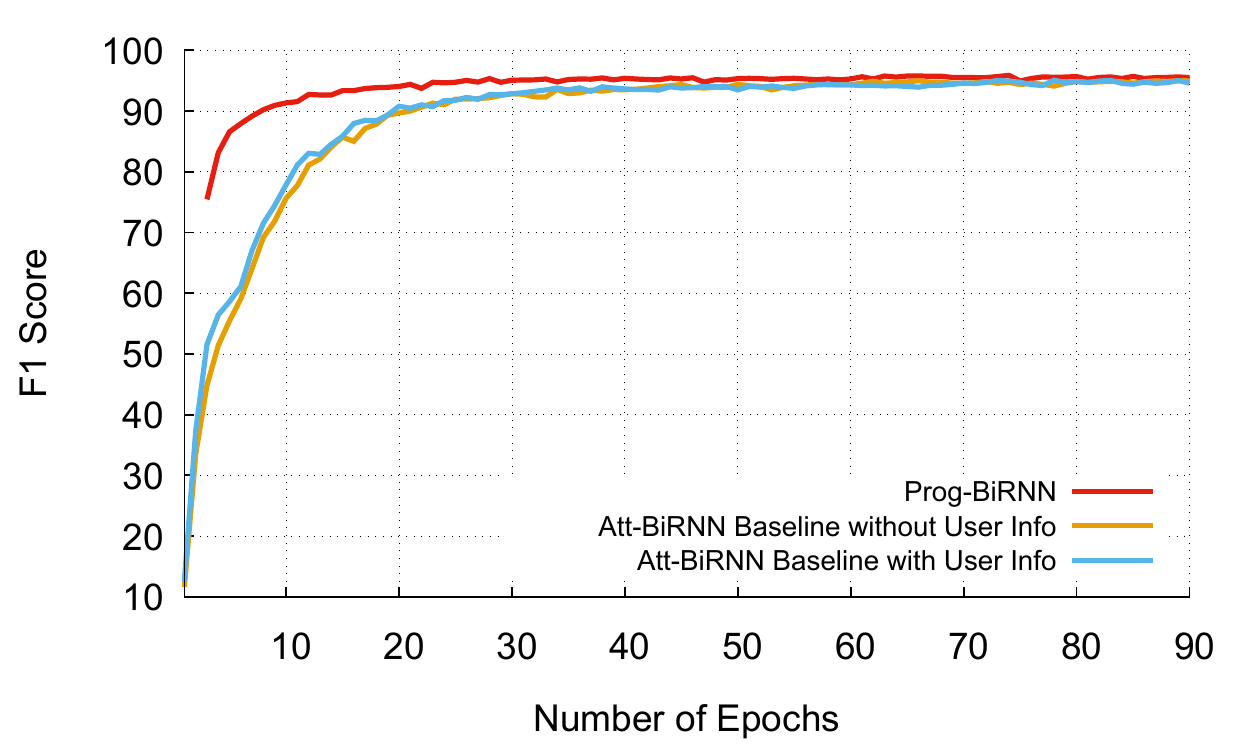}
	\caption{Training time results on full size training set using both contextual location \& preferred time periods as user info}
	\label{fig:training_time}
\end{figure}

We also report the training time between our \kmethod and baseline approaches.
Since our approach mainly focuses on improving slot filling, Figure \ref{fig:training_time} reports the averaged slot filling F1 score after each epoch of training.
Thanks to the small number of user info types, the first user info tagging training phase only takes 3 epochs to achieve over 92\% accuracy, which is sufficient for the second training phase.
As one can see, the number of epochs (3 epochs included) takes to achieve a competitive performance of slot filling is around over 60\% smaller than both two baseline approaches.

\section{Conclusion}\label{sc:discussion}

We present a novel progressive neural network model to train a semantic frame parsing model by incorporating user information.
By using simple user information, we show that our approach not only significantly improves the performance but largely reduces the needs of annotated training set as well.
In addition, our approach also shows its ability to shorten the training time for achieving the competitive performance.
Thus, we enable the quick development of a semantic frame parsing model with less annotated training set in new domains.

%\todo{Our model is independent of the content of user information. Therefore, it can be treated as a personalized parser.}

\newpage
\bibliographystyle{IEEEtran}
\bibliography{yilin,yu,url}

\end{document}